\documentclass{article}

\usepackage[final]{ARLET_2025}

\usepackage[utf8]{inputenc} 
\usepackage[T1]{fontenc}    
\usepackage{hyperref}       
\usepackage{url}            
\usepackage{booktabs}       
\usepackage{amsfonts}       
\usepackage{nicefrac}       
\usepackage{microtype}      
\usepackage{xcolor}         
\usepackage{natbib}  

\usepackage{graphicx}           
\usepackage{subcaption}

\usepackage{amsthm}
\usepackage{algorithm}
\usepackage{algorithmic}
\usepackage{multirow}
\usepackage{amssymb}            
\usepackage{mathtools}          
\usepackage{mathrsfs} 

\newtheorem{theorem}{Theorem}[section]

\theoremstyle{definition}
\newtheorem{definition}[theorem]{Definition}

\theoremstyle{remark}

\title{Human-Inspired Multi-Level Reinforcement Learning}

\author{%
  Mingkang Wu \\
  The University of Texas at San Antonio\\
  \And
  Devin White \\
  Army Educational Outreach Program \\
  \And
  Vernon Lawhern \\
  DEVCOM Army Research Lab \\
  \And
  Nicholas Waytowich \\
  DEVCOM Army Research Lab \\
  \And
  Yongcan Cao \\
  The University of Texas at San Antonio\\
}

\begin{document}

\maketitle

\begin{abstract}


\textcolor{black}{
Reinforcement learning (RL), a common tool in decision making, learns control policies from various experiences based on the associated cumulative return/rewards without treating them differently. Humans, on the contrary, often learn to distinguish from discrete levels of performance and extract the underlying insights/information (beyond reward signals) towards their decision optimization. For instance, when learning to play tennis, a human player does not treat all unsuccessful attempts equally. Missing the ball completely signals a more severe mistake than hitting it out of bounds (although the cumulative rewards can be similar for both cases). Learning effectively from multi-level experiences is essential in human decision making.} 
This motivates us to develop a novel multi-level RL method that learns from multi-level experiences via extracting multi-level information. At the low level of information extraction, we utilized the existing rating-based reinforcement learning~\citep{white2024rating} to infer inherent reward signals that illustrate the value of states or state-action pairs accordingly. At the high level of information extraction, we propose
to extract important directional information from different-level experiences so that policies can be updated towards desired deviation from these different levels of experiences. Specifically, we propose a new policy loss function that penalizes distribution similarities between the current policy and different-level experiences, and assigns different weights to the penalty terms based on the performance levels. Furthermore, the integration of the two levels towards multi-level RL guides the agent toward policy improvements that benefit both reward improvement and policy improvement, hence yielding a similar learning mechanism as humans. To evaluate the effectiveness of the proposed method, we present results for experiments on a few typical environments that show its advantages over the existing rating-based reinforcement learning~\citep{white2024rating}, where a single reward learning was used.
\end{abstract}

\section{Introduction}
\label{sec:introduction}
Recent advancements in reinforcement learning (RL) \citep{sutton1998introduction} have shown promising results in solving complex robotics tasks under the assumption that proper reward functions have been designed~\citep{tang2024deepreinforcementlearningrobotics}. However, in many real world cases, it is often difficult and challenging to define reward functions properly. In these situations, it is often required to include humans users who either provide demonstrations in an offline setting, called learning from demonstrations (LfD), or provide feedback in an online or offline setting, called reinforcement learning from human feedback (RLHF). Popular LfD methods include inverse reinforcement learning (IRL) \citep{argall2009survey} and behavior cloning (BC) \citep{torabi2018behavioral}, which take human expert demonstrations as the input and learn policies that mimic the demonstrations without learning the reward functions. 
In contrast, RLHF takes short video clips or segments and asks humans to provide feedback. The typical form of feedback can take the form of preferences over segment pairs, also known as preference-based reinforcement learning (PbRL) \citep{christiano2017deep}, or ratings for individual segments, also known as rating-based reinforcement learning (RbRL) \citep{white2024rating}. Variants of the PbRL methods include ranking-based RL and crowd-sourcing PbRL \citep{brown2020better, chhan2024crowd}. RLHF methods focus on learning reward functions from the human feedback and then training policies from the learned rewards.

Recent studies in the fields of LfD and RLHF have shown promising capabilities individually. However, jointly learning reward learning and policy learning is challenging in the preference setting since preferences are provided in the relative sense without indicating if a segment is ``optimal" or ``sub-optimal". As a contrary, RbRL utilizes  ratings as the feedback and hence allows the evaluation of individual samples based on their multi-level performance. For example, if a sample is labeled ``Very Good'' (or ``Very Bad'' respectively) which means ``desired'' (or ``undesired'' respectively) that the policy learning should mimic (or distinguish respectively). Meanwhile, the additional ratings of ``Good'', ``Average'', ``Bad'', indicate that the policy learning should deviate it from the three classes in a descending order from humans' cognitive perspective. In other words, the learned policy should be closest to the samples in the ``Very Good'' rating, while less closer to the samples in the ``Good'' rating, and follow a similar trend to be most different from samples in the ``Very Bad'' rating class. Although such a concept is intuitive in humans' decision making process, the current RbRL approach~\citep{white2024rating} only used ratings to learn reward without utilizing different levels of performance behind different rating classes in high-level policy direction learning, which is the focus of the current work. 

In this paper, we propose a novel multi-level reinforcement learning algorithm that leverages human ratings in two levels, namely, low-level reward learning and high-level policy direction learning that guides the agent to systematically deviate from experiences of different rating levels in a human-inspired way. More precisely, the differences from different ratings, or equivalently performance levels, are inverse proportional to their rating classes. Namely, a higher rating class means a lower difference, vice versa. Towards this objective, we first propose a novel Kullback–Leibler (KL) divergence based loss function for different rating classes that penalizes the distribution similarities between the current policy and the segments in different classes. In particular, we classify trajectories not in the highest rating class as failed segments and apply the new loss function with different weights based on their rating labels. For example, when rating class is set to 4 (rating class ``0'', ``1'', ``2'' and ``3''),  the new loss function calculates three distinct KL divergencies between the current policy and trajectories in rating class ``0'', ``1''and ``2'' with different weights. Second, we design this loss function in a modular and flexible manner, allowing it to operate across all rating classes while preserving the original RbRL framework~\citep{white2024rating}, so that the multi-level information derived from the rated data can be applied seamlessly without altering existing training procedures. Third, we conduct experimental studies to evaluate the proposed approach in several environments with different complexity levels and show that the proposed algorithm can yield better performance than the existing RbRL approach.

\section{Related Work}
Learning from demonstrations (LfD) has shown as an effective method to improve RL in dense-reward, sparse-reward and reward-free environments \citep{schaal1996learning, subramanian2016exploration}. Methods, such as DQfD \citep{hester2018deep}, DDPGfD \citep{vecerik2017leveraging} and work in \citep{nair2018overcoming}, have shown significant improvement in RL by employing expert demonstrations to guide the policy searching. Specifically, DQfD and DDPGfD use pre-collected expert demonstrations as replay experiences participating in the policy updating process for Deep Q Network (DQN) and Deep Deterministic Policy Gradient (DDPG) \citep{mnih2013playing, lillicrap2015continuous}. In \citep{nair2018overcoming}, a separate buffer is created to store pre-collect demonstrations which are then used with behavior cloning by penalizing the dissimilarity between the current agent's behavior and the demonstrations. However, expert demonstrations are costly available in practical, which leaves a space for researchers to explore the potential of suboptimal demonstrations or failed experiences in optimizing RL. Noisy demonstrations, with both expert and suboptimal demonstrations, are used as pre-trained data in NAC to initialize a policy and then refine this policy by interacting with environmental rewards \citep{gao2018reinforcement}. 

Failed experiences are used as negative examples to guide the agent's exploration direction in \citep{wu2024offline}. By penalizing the similarity between the current agent's behaviors and those from failed experiences, this method significantly improves offline RL method via overcoming the difficulty in exploration under sparse reward settings. However, failed experiences can differ from each other as they are defined as behaviors that are not optimal \cite{wu2025robust}. Equally treating all of these experiences could lead to unstable exploration.
Existing works such as PbRL \citep{christiano2017deep} and RbRL \citep{white2024rating} label the on-policy segments to indicate their different levels of performance. In reward-free environments, these labels are used to infer reward functions for optimizing RL policies. However, in both PbRL and RbRL, unpreferred segments or those with lower ratings are often abundant, leading to an underutilization of the potential value in failed experiences. To address this, our approach aims to optimize the policy by effectively leveraging failed experiences at different levels.

\section{Preliminaries and Background}
\label{sec:preliminaries}

\subsection{Problem Formulation}
In the context of this paper, we consider a Markov Decision Process (MDP) without reward associated but with ratings, which is defined by a tuple $(\mathcal{S}, \mathcal{A}, P, \gamma, n)$, where $\mathcal{S}$ is the state space, $\mathcal{A}$ the action space, $P$ the state transition probability distribution, $\gamma \in [0, 1)$ is the discount factor that limits the influence of infinite future rewards, and $n$ represents the number of rating classes. At each state $s \in \mathcal{S}$, the RL agent takes an action $a \in \mathcal{A}$, moves to the next state $s'$ determined by $P(s'|s,a)$, where a length-$k$ trajectory $(s_0, a_0, ..., s_{k-1}, a_{k-1})$ is collected to be rated.

In standard RL setting, the environment provides a reward $r: \mathcal{S} \times \mathcal{A} \rightarrow \mathbb{R}$ at each interaction between itself and the RL agent. The goal is to learn a policy $\pi$ that maps states to actions to maximize the expected discounted cumulative rewards. This can be formulated by the state-action value function
\begin{equation}\label{Qnetwork}
    Q(s,a) = \mathbb{E}_{a_t\sim\pi}\left[\sum_{t=0}^{\infty} \gamma^tR(s_t,a_t) \right],
\end{equation}
where $t$ represents the $t$\textsuperscript{th} timestep in the training process. The performance of a policy $\pi$ is normally evaluated by the discounted cumulative rewards
\begin{equation}
    J(\theta) = \mathbb{E}_{s\sim\mu}\left[Q(s,\pi(s|\theta))\right],
\end{equation}
where $\mu$ represents the initial state distribution and $\theta$ is the policy network parameter. \textcolor{black}{The policy $\pi$ defines the agent's behavior by specifying the probability distribution over actions given the current state. The goal of an RL agent is to find an optimal policy $\pi^*$ that maximizes the expected cumulative reward over time, \textit{i.e.},}
$
    \pi^* = \arg\max_{\pi} \mathbb{E}_{\pi} \left[ \sum_{t=0}^{T} \gamma^t r_t \right],
$
where $r_t$ is the reward received at time step $t$ and $T$ is the time horizon. 

Note that the update of policy relies much on the cumulative rewards, which implies the existence of rewards in each state-action pair, also referred to dense reward environments. However, in reward-free environments where rewards are not present, the standard RL methods fail to work due to the lack of (reward) knowledge to guide the policy search. 

\begin{figure*}
    \centering
    \includegraphics[width=0.9\linewidth]{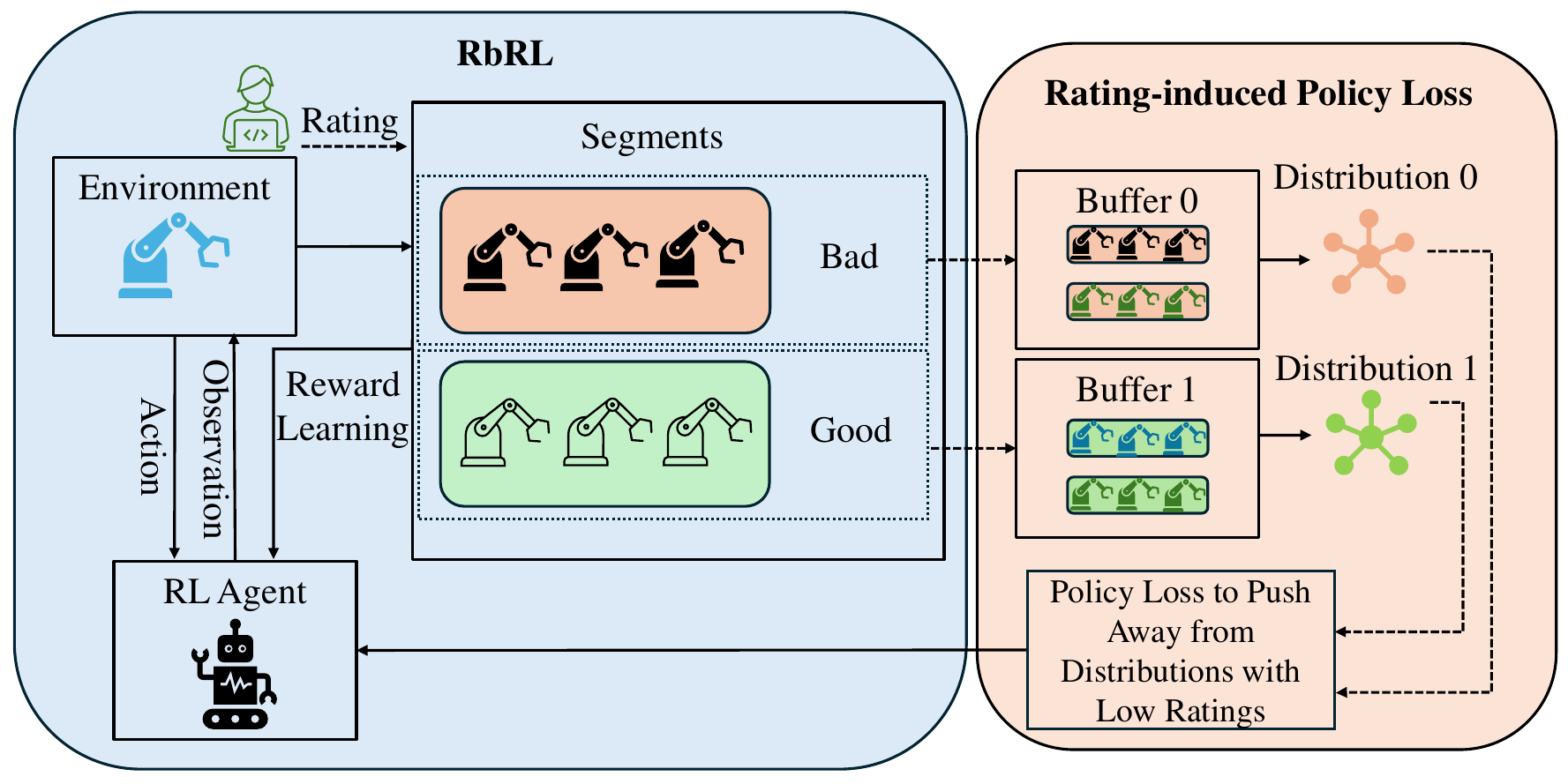}
    \caption{A schematic illustration of the proposed algorithm and its relationship with RbRL.}
    \label{fig:flowchart}
\end{figure*}

\subsection{Rating-Based Reinforcement Learning}

Due to the lack of rewards, the rating-based reinforcement learning (RbRL) \citep{white2024rating} learns a reward model $\hat{r}: \mathcal{S} \times \mathcal{A} \rightarrow \mathbb{R}$ that predicts reward $\hat{r}(s, a)$ for each state-action pair during interaction. Given a length-\textit{j} segment $\sigma = (s_1, a_1, ..., s_j, a_j)$, the cumulative discounted reward $\hat{R}(\sigma) := \sum^j_t=1\gamma^{t-1}\hat{r}(s_t, a_t)$ based on $\hat{r}$ provides an estimated cumulative reward. 

The key idea of RbRL is to train a reward model $\hat{r}:(s,a)\mapsto R$ that can explain why the existing samples were given their corresponding ratings. First, the cumulative predicted reward is normalized across a batch $\tilde{R}(\sigma) = \frac{\hat{R}(\sigma) - \min_{\sigma' \in X} \hat{R}(\sigma')} {\max_{\sigma' \in X}\hat{R}(\sigma') - \min_{\sigma' \in X} \hat{R}(\sigma')}$. Then, the probability of each sample in each rating class was computed as
\begin{equation}\label{eq:Qfun_old}
     Q_{\sigma}(i) = \frac{e^{-k(\tilde{R}(\sigma)-\bar{R}_i)(\tilde{R}(\sigma)-\bar{R}_{i + 1})}}{\sum_{j=0}^{n-1} e^{-k(\tilde{R}(\sigma)-\bar{R}_j)(\tilde{R}(\sigma)-\bar{R}_{j + 1})}},
\end{equation}
where $\bar{R}_i$ and $\bar{R}_{i+1}$ are the lower and upper bound value for the $i\textsuperscript{th}$ class respectively. The reward predictor $\hat{r}$ was trained by minimizing the cross-entropy loss given by  
\begin{equation}\label{eq:crossentropy}
    L(\hat{r}) = -\sum_{\sigma\in X} \left( \sum_{i=0}^{n-1}\mu_{\sigma}(i) \log \big(Q_{\sigma}(i)\big) \right),
\end{equation}
where $X$ is the set containing all samples, and $\mu_{\sigma}(i)=1$ if the sample is labeled in the class $i$ and $\mu_{\sigma}(i)=0$ otherwise.

Once the reward model $\hat{r}$ was learned, one can use any existing RL algorithm, such as PPO, DDPG, SAC, to train a control policy. Note that the rated samples in this method were only used in reward learning without investigating their value on policy learning directly. For example, it is intuitive that a good policy should deviate more from samples with lower ratings while less from samples with higher ratings. Hence, the rated samples can provide additional values in direct policy shaping via designing an appropriate loss function, which is the focus of the next section.   

\section{Multi-Level Reinforcement Learning}
\label{sec:proposed method}

Figure \ref{fig:flowchart} shows the schematic structure of the new multi-level reinforcement learning approach. Its loss function can be divided into two components. The first component is the classic loss function for gradient-based RL algorithms based on the learned reward from the rating-based reinforcement learning algorithm~\citep{white2024rating}. The second component is a new loss that penalizes the similarities between the samples from the current policies and the samples from different rating classes. Since different rating classes contain samples with different performance levels, the weights used in the penalty should be different for different rating classes. The main idea of the new loss component is to use the Kullback–Leibler (KL) divergence to quantify the similarities (equivalently, differences) with an descending weight for rating classes from low to high, which ensures more difference from low rating classes, namely, low performance levels, and less difference from high rating classes, namely, high performance levels. It is worth emphasizing that samples with high performance levels are still undesired but show better performance than those with low performance levels. Hence, all samples used in the new loss are considered failure, but with different performance levels characterized by different rating classes.

\subsection{Overall Loss Function}
The proposed new loss function is given by
\begin{equation}\label{eq:newloss}
    \nabla_{\theta}J(\pi_{\theta}) = \mathbb{E}_{\pi_{\theta}}[\nabla_{\theta}\log(\pi_{\theta})\hat{R}(\sigma_\theta)] - \nabla_{\theta}\sum_{i=0}^{n-2}\omega_i D_{KL}(D_i \parallel D_{\pi_\theta}),
\end{equation}
where the first component is the classic loss based on the learned reward and the second component is the new loss, $\sigma_\theta$ includes the trajectories sampled from the current policy at each training batch, $D_{KL}$ denotes the KL divergence between two different distributions (please refer to Definition \ref{pro: multivariate guassian distribution} below), $\omega_i$ is the weight applied to the KL divergence in a descending order as rating level moves from lowest to highest, while $D_i$ and $D_{\pi_{\theta}}$ represent the trajectory distributions of rating class $i$ and the current policy, respectively. This new policy gradient loss effectively penalizes the similarities between the current policy and different rating levels of the failed experiences such that the policy is updated in the direction of continuously improving the performance thanks to the descending order for the weight $\omega_i$. 

To facilitate the understanding of this new loss, we will now provide some details. In the context of the new loss function, we utilize all, except the highest, rating classes. For example, consider the case of 4 rating classes, namely ``very bad", ``bad", ``good" and ``very good", whose ratings are ``0", ``1", ``2" and ``3", respectively. The segments rated as ``very bad" are stored in buffer ``0", namely, $R_0$. The segments rated as ``bad" in buffer ``1", namely, $R_1$. The segments rated as ``good" in buffer ``2", namely $R_2$. Finally, the segments rated as ``very good" in buffer ``3", namely, $R_3$. All trajectories in $R_0$, $R_1$ and $R_2$ are used in~\eqref{eq:newloss}. In other words, the rating classes used in~\eqref{eq:newloss} include class $0$ to class $n-2$, which is reflected in the second term of~\eqref{eq:newloss} with the summation computed for $i=0$ to $i=2$. 

In RL, a policy is initialized with a probability density function, also referred to as a distribution, mapping from the states input and the actions output. The main goal of the RL agent is to learn an optimal policy that maximizes the cumulative rewards. In the context of rating-based reinforcement learning, the cumulative rewards are based on the estimated rewards since the reward is unknown or does not exist. Besides the class loss term, the proposed new loss term penalizes the deviation between the current policy and different rating levels of failed examples. To this end, we employ the multivariate Gaussian distribution \citep{goodman1963statistical} to represent trajectories in each rating class, and use KL divergence \citep{hershey2007approximating} to measure the distribution-wise difference between the current policy and each failure distribution. The main motivation behind this new loss term is to provide a direct policy shaping mechanism towards exploring areas that are different from various rating levels of the failed samples at different penalty weights. In other words, the RL agent will be pushed away from samples and its neighboring regions at different performance levels. The lower the performance level is, the larger the push-away force will be applied.  

\subsection{Policy Loss Based on KL divergence}

Since the purpose of our approach is to quantify the deviation between the current policy and different rating levels of failed samples via computing their KL divergencies, we employ  multivariate Gaussian distribution to represent the distributions by computing the essential components, including covariance matrix and mean values, of trajectories in low rating classes and those sampled from the current policy. Specifically, we propose to compute the KL divergence between the current policy and the samples in different rating levels of failed samples as follows. 

\begin{definition} \label{pro: multivariate guassian distribution}
    For any two distributions, $P$ and $Q$, parameterized by their means $\mu_p$ and $\mu_q$ and covariance $\Sigma_p$ and $\Sigma_q$. Mathematically, the KL divergence between $P(\mu_p, \Sigma_p)$ and $Q(\mu_q, \Sigma_q)$ is
    \begin{equation} \label{eq: KL}
        D_{KL}(P\parallel Q) = \frac{1}{2}(\mathrm{Tr}(\Sigma^{-1}_q\Sigma_p) + (\mu_p - \mu_q)^T\Sigma^{-1}_q(\mu_q - \mu_p) - k + \ln(\frac{\det(\Sigma_q)}{\det(\Sigma_p)})),
    \end{equation}
    where, $\mathrm{Tr}(\cdot)$ is the trace of a given matrix, $\det(\cdot)$ represents the determination of a given matrix, and $k$ is the number of features. 
\end{definition}

Definition~\ref{pro: multivariate guassian distribution} provides a measure of how different the two distributions $P$ and $Q$ are. For example, we have a set of failed trajectories $\sigma_i = ((s_{i_0}^0, a_{i_0}^0, s_{i_0}^1, a_{i_0}^1, ...), ..., (s_{i_m}^0, a_{i_m}^0, s_{i_m}^1, a_{i_m}^1, ...))$ in rating class \textit{i} containing $m$ trajectories, where $i$ is not the highest rating class. Another set of trajectories $\sigma_{\pi_\theta} = ((s_{\pi}^0, a_{\pi}^0, s_{\pi}^1, a_{\pi}^1, ...), ..., (s_{\pi}^0, a_{\pi}^0, s_{\pi}^1, a_{\pi}^1, ...))$ sampled from the current policy $\pi$ parameterized by $\theta$ represents the behaviors of the RL agent in the current training batch. According to Definition \ref{pro: multivariate guassian distribution}, the KL divergency between the distribution of rating class \textit{i} and the distribution of the current policy can be formulated as
\begin{equation} \label{eq: dis between i&pi}
   D_{KL}(D_i \parallel D_\pi) = \frac{1}{2}(\mathrm{Tr}(\Sigma^{-1}_{D_\pi}\Sigma_{D_i})+ (\mu_{D_i} - \mu_{D_\pi})^T\Sigma^{-1}_\pi(\mu_\pi - \mu_{D_i})+ \ln(\frac{\det(\Sigma_{D_\pi})}{\det(\Sigma_{D_i})})). 
\end{equation}
Since $D_{KL}(D_i \parallel D_\pi)$ is used to compute the policy gradient in \eqref{eq:newloss}, the constant $k$ in \eqref{eq: KL} can be omitted here.

Consider the case of 4 different rating classes (``0", ``1", ``2" and ``3"). Following the computation of the KL divergence between the current policy and the samples in rating classes 0, 1, and 2 (except class 3, which is the highest rating class and hence not included as explained in the Subsection ``Overall Loss Function''), the overall loss function can be written as
\begin{align}
    \nabla_{\theta}J(\pi_{\theta}) = & \mathbb{E}_{\pi_{\theta}}[\nabla_{\theta}\log(\pi_{\theta})\hat{R}(\sigma_\theta)] \\
    & - \left[\omega_0 D_{KL}(D_0 \parallel D_{\pi_\theta}) + \omega_1 D_{KL}(D_1 \parallel D_{\pi_\theta}) \notag + \omega_2 D_{KL}(D_2 \parallel D_{\pi_\theta})\right],
\end{align}

where $\hat{R}(\sigma_\theta)$ represents the cumulative predicted rewards of the trajectories sampled by the current policy $\pi_\theta$, $\omega_0$, $\omega_1$ and $\omega_2$ represents the weights for KL divergencies between rating classes ``0'', ``1'' and ``2'' and the current policy $\pi_{\theta}$, respectively. The weights are assigned in the descending order, namely $\omega_0>\omega_1>\omega_2$, such that current policy $\pi_\theta$ is pushed away from the distribution of rating class ``0" more than distributions of rating classes ``1'' and ``2''. 
The pseudocode of the proposed new approach is given in Algorithm \ref{alg: pseudo code}.

\begin{algorithm}[tb]
   \caption{} 
   \label{alg: pseudo code}
\begin{algorithmic}
   \STATE {\bfseries Input:} rating classes $n$, rating buffers $R_1, ..., R_n$, initial reward predictor $\hat{r}$, KL divergency weight $\omega_i$, total training cycles $T$, total training cycle $M$ for $\hat{r}$
   
   \STATE Initialize RL policy $\pi_{\theta_0}$
   \FOR{$i=1$ {\bfseries to} $M$}
   \STATE Sample trajectories $\sigma$ from $\pi_{\theta_0}$
   \STATE Rate $\sigma$ by humans
   \STATE Update $\hat{r}$ based on human ratings
   \ENDFOR

   \FOR{$i=1$ {\bfseries to} $T$}
   \STATE Extract trajectories $\sigma_1, .., \sigma_n$ from rating buffers $R_1, ..., R_n$
   \STATE Update policy with $\hat{r}$

    \begin{equation}
    \nabla_{\theta}J(\pi_{\theta})\notag = \mathbb{E}_{\pi_{\theta}}[\nabla_{\theta}\log(\pi_{\theta})\hat{R}(\sigma_\theta)]- \nabla_{\theta}\sum_{i=0}^{n-2}\omega_i D_{KL}(D_i \parallel D_{\pi_\theta})\nonumber
   \end{equation}

   \STATE Update policy parameter $\theta_i$

   \begin{equation}
   \theta_{i+1} \leftarrow \theta_{i} + \alpha\nabla_{\theta_{i}} J(\pi_{\theta_{i}})\nonumber 
   \end{equation}

   \ENDFOR

\end{algorithmic}
\end{algorithm}

\section{Experiments and Results}\label{sec: experiment and result}

To evaluate the effectiveness of our proposed method, we compare the new method, labeled as RbRL-KL, with RbRL across 6 DeepMind Control environments \citep{tassa2018deepmind}, namely, Cartpole-balance, Ball-in-cup, Finger-spin, HalfCheetah, Walker and Quadruped. These environments are characterized with continuous state and action spaces, and each vary in complexity. Specifically, Cartpole-balance is characterized by a simple 5-dimensional state space for cart and pole dynamics, and a 1-dimensional action space, representing discrete control forces. Ball-in-cup has an 8-dimensional state space capturing the relative positions and velocities of the involved objects, and a 2-dimensional action space controlling the cup's motion. Finger-spin is characterized with a 12-dimensional state space representing the finger and object dynamics, and a 2-dimensional action space controlling the finger's movement. HalfCheetah has a 17-dimensional state space, capturing joint positions, velocities, and body orientation, and a 6-dimensional action space, which corresponds to the torques applied to each joint. Walker has a 24-dimensional state space representing joint angles, velocities, and torso orientation, and a 6-dimensional action space, controlling forces applied to each limb. Quadruped is a more complex environment which characterized with a 78-dimensional state space, including joint positions, velocities, and full body orientation, along with a 12-dimensional action space, enabling torque-based control over each leg joint. Among these environments, Cartpole-balance is the simplest environment while Quadruped is the most complex one. We focus on evaluating and understanding how RbRL-KL performs across different levels of complexity. While these environments include built-in reward functions, we deliberately avoid using them to preserve a reward-free testing setup. Instead, the original reward functions are used to purely evaluate the performance of the trained policies.

\begin{figure*}[!ht]
\centering
    \begin{subfigure}{0.3\textwidth}
        \centering
        \includegraphics[width=1.1\textwidth]{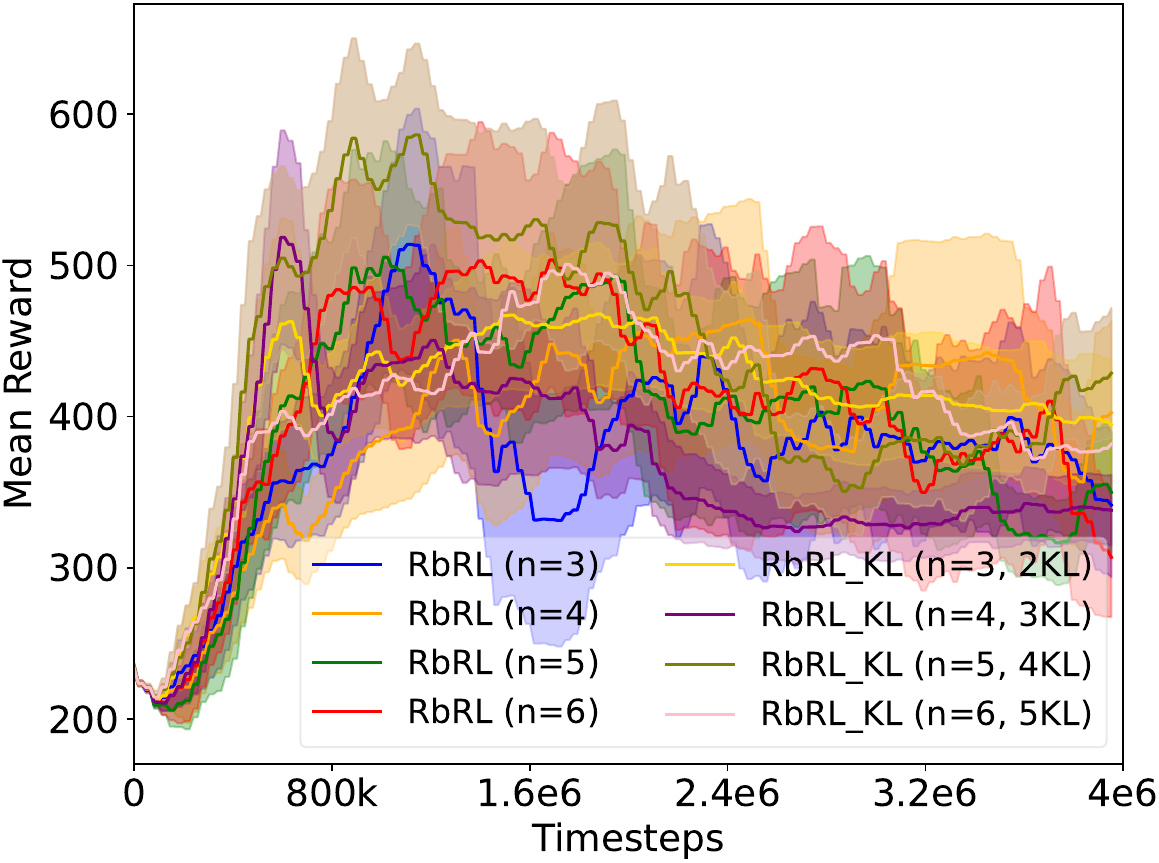}
        \caption{ {Cartpole-balance}}
        \label{fig:cartpole_balance}
    \end{subfigure}
    \hfill
    \begin{subfigure}{0.3\textwidth}
        \centering
        \includegraphics[width=1.1\textwidth]{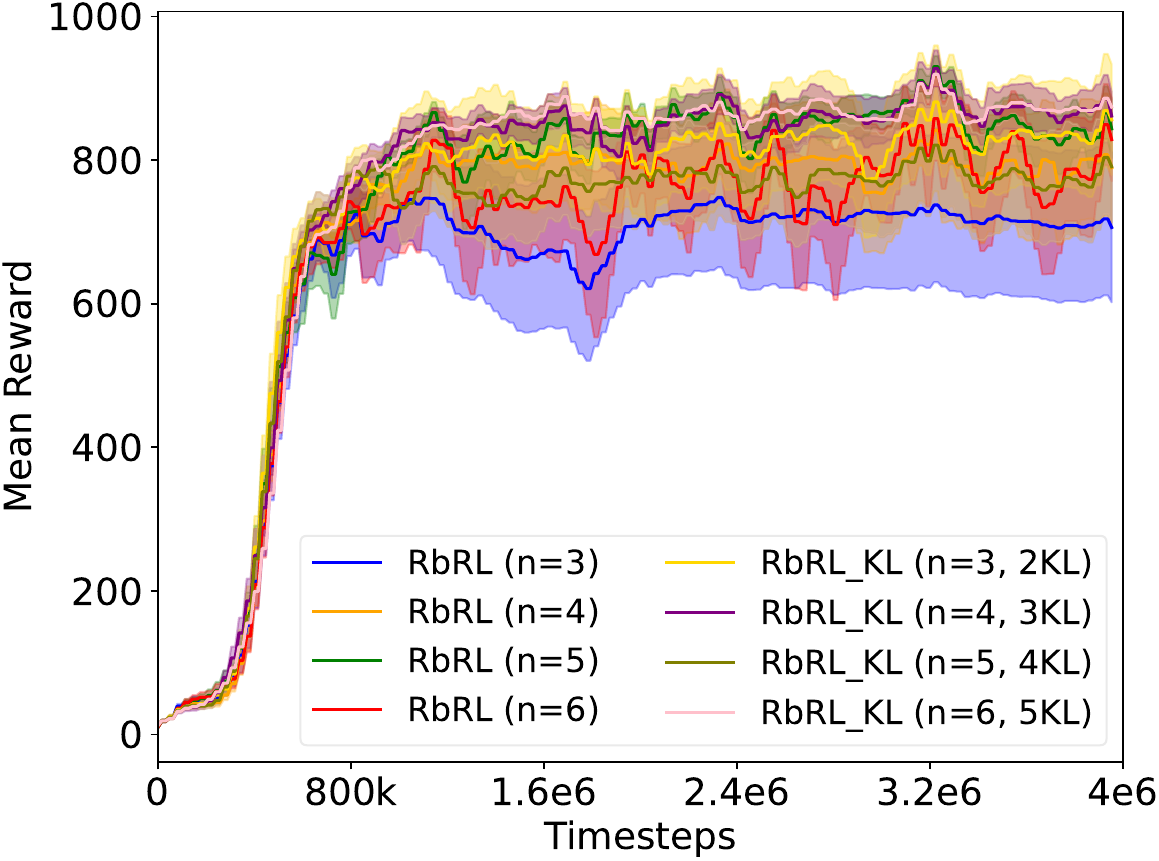}
        \caption{ {Ball-in-cup}}
        \label{fig:ball_in_cup}
    \end{subfigure}
    \hfill
    \begin{subfigure}{0.3\textwidth}
        \centering
        \includegraphics[width=1.1\textwidth]{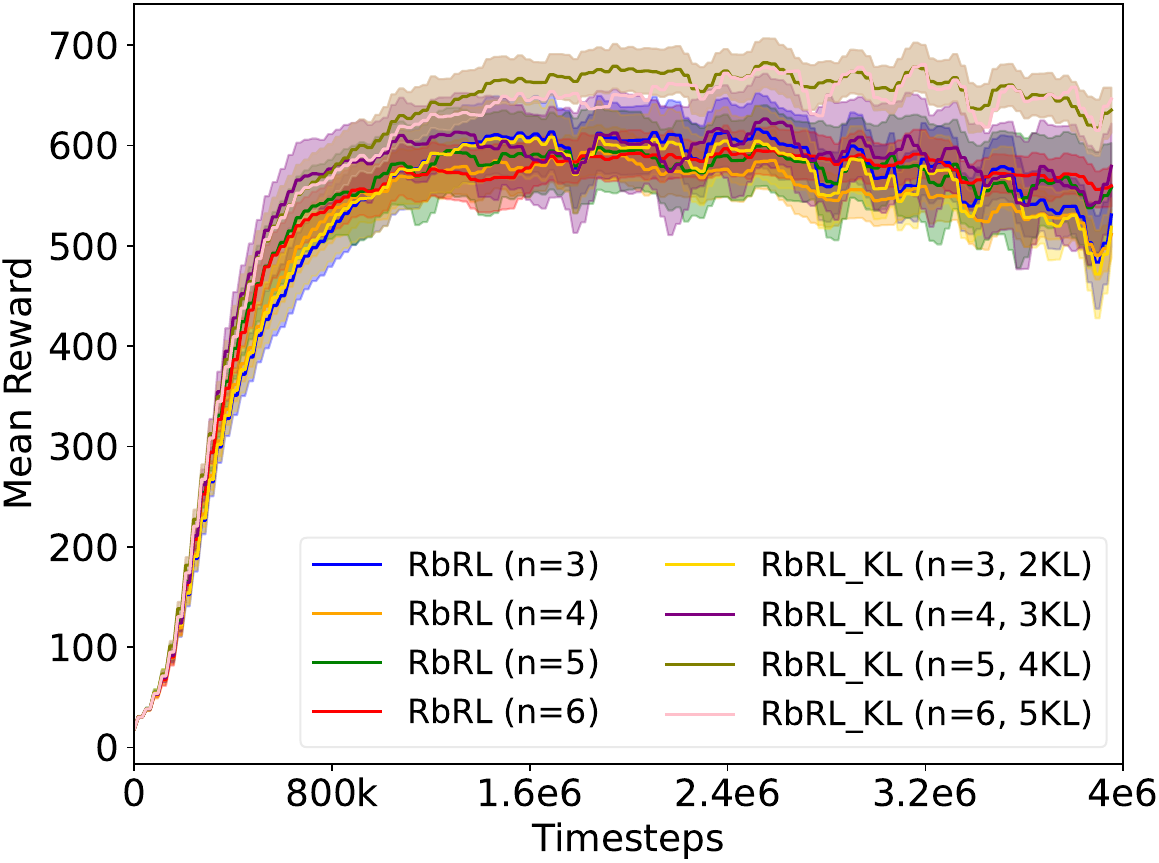}
        \caption{Finger-spin}
        \label{fig:finger_spin}
    \end{subfigure}
    \hfill
    \begin{subfigure}{0.29\textwidth}
        \centering
        \includegraphics[width=1.1\textwidth]{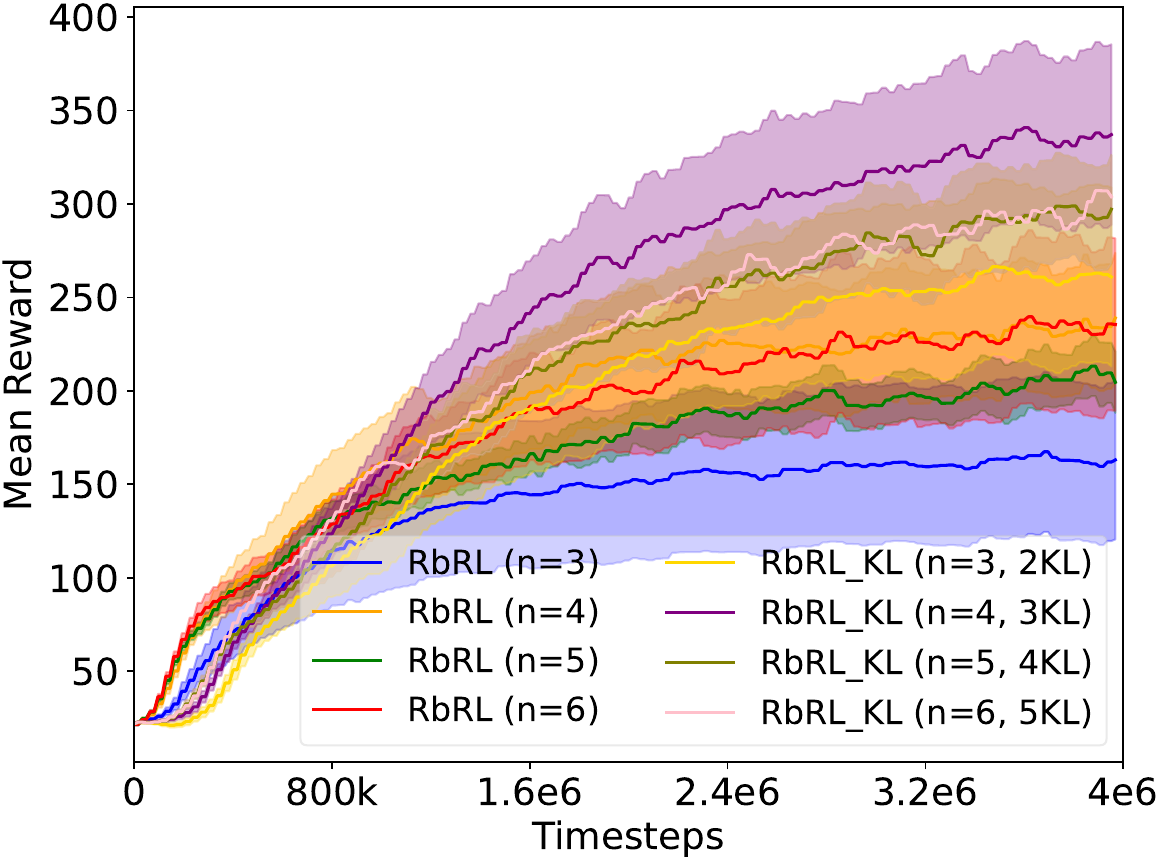}
        \caption{ {HalfCheetah}}
        \label{fig:half}
    \end{subfigure}
    \hfill
    \begin{subfigure}{0.3\textwidth}
        \centering
        \includegraphics[width=1.1\textwidth]{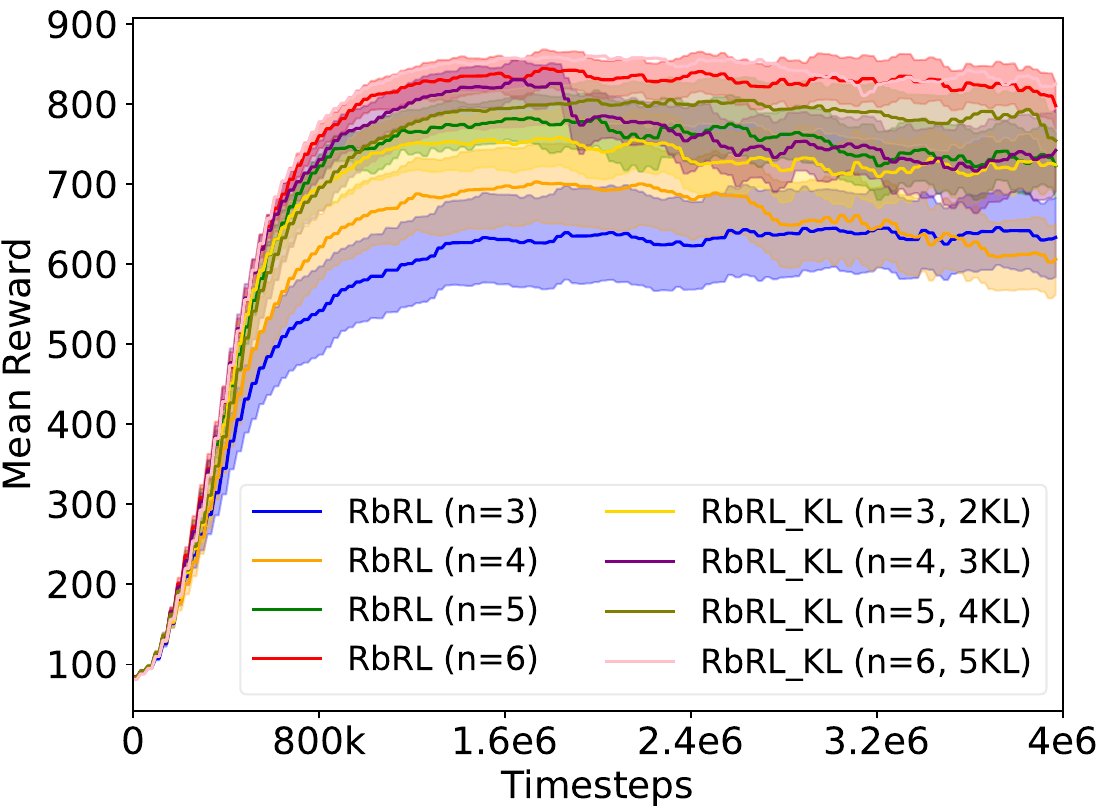}
        \caption{ {Walker}}
        \label{fig:walker}
    \end{subfigure}
    \hfill
    \begin{subfigure}{0.3\textwidth}
        \centering
        \includegraphics[width=1.1\textwidth]{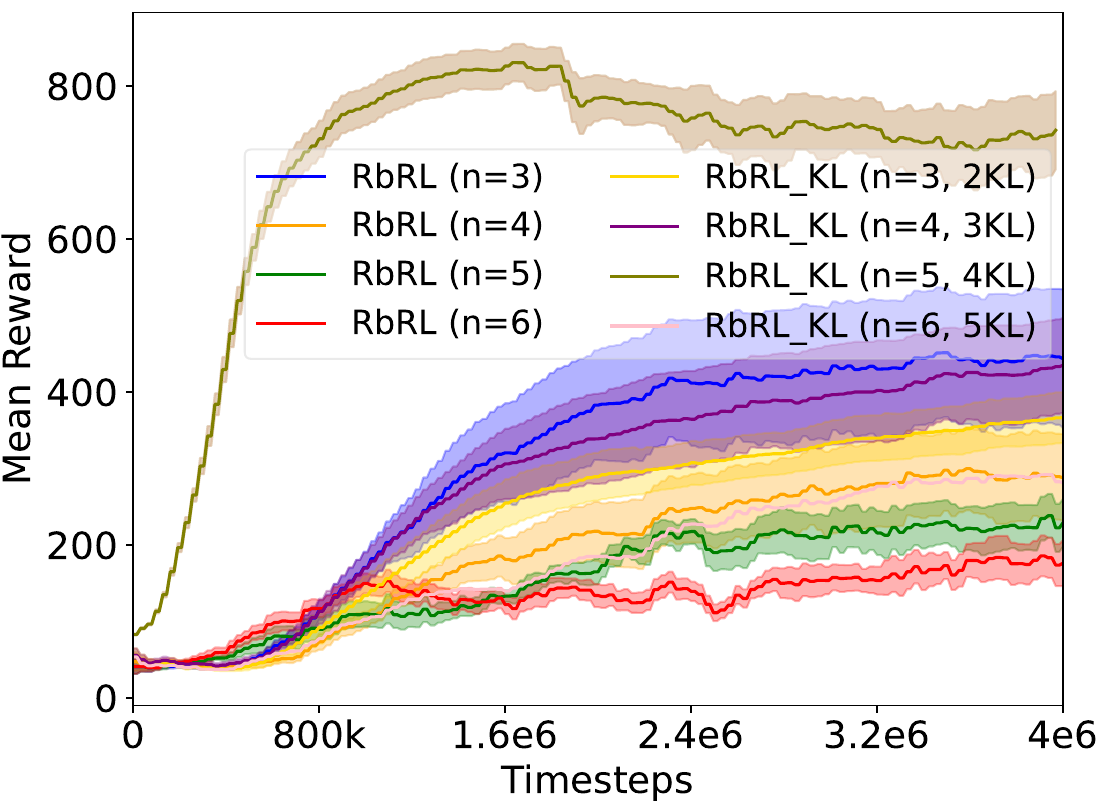}
        \caption{ {Quadruped}}
        \label{fig:quad}
    \end{subfigure}
    \caption{Learning curves of different algorithms across six environments. The plots show the mean (solid line) with the standard error (shaded area) over 10 runs.
}\label{fig:results}
\end{figure*}

\begin{table*}[h!]
  \centering
  \caption{Hyperparameters used in experiments.}
  \renewcommand{\arraystretch}{1.2}
  \resizebox{\textwidth}{!}{ 
    \begin{tabular}{c|c|c|c|c|c|c|c|c|c}
      \hline
      Environment & Clip Param $\epsilon$ & Learning Rate $\alpha$ & Batch Size & Hidden Layers & $\omega_0$ & $\omega_1$ & $\omega_2$ & $\omega_3$ & $\omega_4$ \\
      \hline
      Cartpole-balance & 0.4 & 0.00005& 128 & 3 & 1.0 & 0.5 & 0.25 & 0.125 & 0.06 \\
      Ball-in-cup & 0.4 & 0.00005& 128 & 3 & 1.0 & 0.5 & 0.25 & 0.125 & 0.06 \\
      Finger-spin & 0.4 & 0.00005& 128 & 3 & 1.0 & 0.5 & 0.25 & 0.125 & 0.06 \\
      HalfCheetah & 0.4 & 0.00005 & 128 & 3 & 1.0 & 0.5 & 0.25 & 0.125 & 0.06 \\
      Walker & 0.4 & 0.00005 & 128 & 3 & 1.0 & 0.5 & 0.25 & 0.125 & 0.06 \\
      Quadruped & 0.4 & 0.00005& 128 & 3 & 1.0 & 0.5 & 0.25 & 0.125 & 0.06 \\
      \hline
    \end{tabular}
  }
  
  \label{tab:parameters}
\end{table*}

To further evaluate the effectiveness of our proposed method, we compare RbRL-KL against RbRL across rating classes of 3, 4, 5, and 6, corresponding to training with 2, 3, 4, and 5 KL divergence terms, respectively. The hyperparameters used in our experiments are provided in Table \ref{tab:parameters}. To ensure reproducibility, each setting is run 10 times using different random seeds. Table \ref{tab:results} presents the average cumulative rewards with standard errors over 10 runs, and Figure \ref{fig:results} compares the learning curves of both algorithms across different rating classes. The results show that RbRL-KL outperforms RbRL in most environments. In particular, RbRL-KL consistently achieves better performance on Cartpole-balance, Ball-in-cup, HalfCheetah, and Walker, while it only underperforms in Finger-spin and Quadruped when $n=3$. It is worth noting that RbRL outperforms RbRL-KL under the lower rating class because the lower-rated segments form a broad, undifferentiated group. As a result, the KL-divergence terms in RbRL-KL apply a more uniform deviation across all these segments, reducing their impact on policy learning. Therefore, the trained policy is more likely to achieve the local optima. Instead, a more refined rating system will provide better sample separation, allowing the KL-based policy loss in RbRL-KL to learn more effectively. Apart from these specific cases, the proposed RbRL-KL consistently outperforms RbRL, demonstrating its superior performance. To provide a clearer view of the algorithm’s advantages, Table \ref{tab:percentage_improvement} illustrates the percentage improvement achieved by RbRL-KL over conventional reward learning from human ratings. Positive values indicate performance enhancement, showing that RbRL-KL generally improves learning, while minor decreases appear only in Finger-Spin and Quadruped under the low rating class.

\begin{table*}[ht!]
  \centering
  \caption{Empirical return comparison among different algorithms.}
  \begin{tabular}{c|c|c|c|c}
    \hline
    \multirow{2}{*}{\shortstack{Environment\\($\uparrow$ complexity)}} & \multicolumn{4}{c}{Empirical Return} \\
    \cline{2-5}
    & RbRL (n=3) & RbRL (n=4) & RbRL (n=5) & RbRL (n=6) \\
    \hline
    Cartpole-balance & $341.50 \pm 47.77$ & $402.55 \pm 60.65$ & $349.84 \pm 39.95$ & $306.92 \pm 39.55$ \\
    Ball-in-cup & $706.20 \pm 104.26$ & $789.30 \pm 84.62$ & $698.15 \pm 104.55 $ & $828.62 \pm 57.98$ \\
    Finger-spin & $\mathbf{530.79 \pm 31.15}$ & $511.55 \pm 24.25$ & $557.88 \pm 44.52$ & $559.73 \pm 20.49$ \\
    HalfCheetah & $163.02 \pm 42.49$ & $ 238.99 \pm 34.81$ & $204.59 \pm 16.37$ & $235.46 \pm 46.10$ \\
    Walker & $633.34 \pm 49.35$ & $606.14 \pm 44.10$ & $722.94 \pm 40.81$ & $797.90 \pm 28.94$ \\
    Quadruped & $\mathbf{454.39 \pm 89.44}$ & $308.48 \pm 62.29$ & $227.52 \pm 39.92$ & $199.83 \pm 45.71$ \\
    
    \hline


    & RbRL-KL (n=3) & RbRL-KL (n=4) & RbRL-KL (n=5) & RbRL-KL (n=6) \\
    \hline
    Cartpole-balance & $\mathbf{394.62 \pm 41.30}$ & $\mathbf{417.54 \pm 58.89}$ & $\mathbf{428.64 \pm 43.10}$ & $\mathbf{381.79 \pm 33.91}$ \\
    Ball-in-cup & $\mathbf{856.25 \pm 77.26}$ & $\mathbf{861.47 \pm 33.34}$ & $\mathbf{790.94 \pm 78.00}$ & $\mathbf{873.92 \pm 16.03}$ \\
    Finger-spin & $518.51 \pm 29.03$ & $\mathbf{579.27 \pm 42.62}$ & $\mathbf{635.18 \pm 22.44}$ & $\mathbf{646.37 \pm 20.13}$ \\
    HalfCheetah & $\mathbf{260.93 \pm 47.89}$ & $\mathbf{337.04 \pm 48.27}$ & $\mathbf{297.11 \pm 29.02}$ & $\mathbf{303.88 \pm 38.07}$ \\
    Walker & $\mathbf{724.21 \pm 31.04}$ & $\mathbf{742.05 \pm 50.99}$ & $\mathbf{754.35 \pm 32.79}$ & $ \mathbf{825.18 \pm 26.44}$\\
    Quadruped & $420.27 \pm 44.00$ & $\mathbf{477.29 \pm 69.89}$ & $\mathbf{742.05 \pm 50.99}$ & $\mathbf{306.78 \pm 75.69}$ \\
    \hline
  \end{tabular}

  \label{tab:results}
\end{table*}

\begin{table*}[h!]
\centering
\caption{Percentage improvement of RbRL-KL over RbRL across different environments and values of $n$.}
\begin{tabular}{l|r|r|r|r}
\hline
\textbf{Environment} & \textbf{n=3 (\%)} & \textbf{n=4 (\%)} & \textbf{n=5 (\%)} & \textbf{n=6 (\%)} \\
\hline
Cartpole-balance & $\mathbf{15.54}$ & $\mathbf{3.72}$ & $\mathbf{22.50}$ & $\mathbf{24.37}$ \\
Ball-in-cup      & $\mathbf{21.27}$ & $\mathbf{9.14}$   & $\mathbf{13.30}$ & $\mathbf{5.47}$  \\
Finger-spin      & $-2.32$ & $\mathbf{13.24}$  & $\mathbf{13.85}$ & $\mathbf{15.47}$ \\
HalfCheetah      & $\mathbf{60.03}$ & $\mathbf{40.99}$  & $\mathbf{45.24}$ & $\mathbf{29.06}$ \\
Walker           & $\mathbf{14.35}$ & $\mathbf{22.39}$  & $\mathbf{4.34}$  & $\mathbf{3.42}$  \\
Quadruped        & $-7.51$ & $\mathbf{54.74}$  & $\mathbf{225.98}$ & $\mathbf{53.46}$ \\
\hline
\end{tabular}

\label{tab:percentage_improvement}
\end{table*}


\section{Limitations and Future Work}

One of the limitations of the proposed approach is the variability in individual rating standards, which may be noisy due to differences in how participants interpret and evaluate the environment. Since RbRL relies on users having a basic understanding of the task and environment, these inconsistencies may reduce the reliability of the collected individual ratings. To address this, future work will involve designing a crowdsourcing framework to gather a diverse range of ratings from various participants. Developing effective noise filtering methods is important to ensure the collected data is more representative and robust, improving the reliability and accuracy of the learned models.

Although $\omega_i$ should be selected in a descending order with respect to the rating level, it remains an open question how to select the specific value of $\omega_i$. Ablation studies will be needed to optimize the performance by selecting the best $\omega_i$.
Another interesting future research is to integrate the new KL loss terms with other RL methods in dense-reward, sparse-reward, and reward-free settings to further evaluate its effectiveness since the KL loss terms can be implemented with/without rewards.  Finally, it is important to develop a systematic approach that uncovers the right number of rating classes for a given environment since a larger number of rating classes may not always lead to better performance. One potential idea is to bring the idea of Just Noticeable Difference, which quantifies the smallest change in a stimulus to be observed by humans, in psychology to determine the boundaries between different ratings. We will explore these directions as parts of our future work.


\bibliography{main}
\bibliographystyle{plainnat}

\appendix


\end{document}